\pdfoutput=1

\documentclass[11pt]{article}

\usepackage[final]{acl}

\usepackage{times}
\usepackage{latexsym}

\usepackage[T1]{fontenc}

\usepackage[utf8]{inputenc}

\usepackage{microtype}

\usepackage{makecell}
\usepackage{graphicx}
\usepackage{appendix}
\usepackage{multirow}
\usepackage{amsfonts,amssymb}

\newcommand{\MODELNAME}{GAIN}
\newcommand{\FULLMODELNAME}{gazetteer-adapted integration network}

\title{USTC-NELSLIP at SemEval-2022 Task 11: Gazetteer-Adapted Integration Network for Multilingual Complex Named Entity Recognition}

\author{Beiduo Chen\textsuperscript{1}, Jun-Yu Ma\textsuperscript{1}, Jiajun Qi\textsuperscript{1}, Wu Guo\textsuperscript{1}, Zhen-Hua Ling\textsuperscript{1} \and Quan Liu\textsuperscript{2}\\
\textsuperscript{1}National Engineering Laboratory for Speech and Language Information Processing,\\
University of Science and Technology of China\\ \textsuperscript{2}State Key Laboratory of Cognitive Intelligence, iFLYTEK Research \\
\texttt{\{beiduo,mjy1999,jiajun97\}@mail.ustc.edu.cn} \\
\texttt{\{guowu,zhling\}@ustc.edu.cn, quanliu@iflytek.com}}

\begin{document}

\maketitle

\begin{abstract}
This paper describes the system developed by the USTC-NELSLIP team for SemEval-2022 Task 11 Multilingual Complex Named Entity Recognition (MultiCoNER).
We propose a {\FULLMODELNAME{}} (\MODELNAME{}) to improve the performance of language models for recognizing complex named entities.
The method first adapts the representations of gazetteer networks to those of language models by minimizing the KL divergence between them.
After adaptation, these two networks are then integrated for backend supervised named entity recognition (NER) training.
The proposed method is applied to several state-of-the-art Transformer-based NER models with a gazetteer built from Wikidata, and shows great generalization ability across them. 
The final predictions are derived from an ensemble of these trained models.
Experimental results and detailed analysis verify the effectiveness of the proposed method.
The official results show that our system ranked \textbf{1st} on three tracks (Chinese, Code-mixed and Bangla) and \textbf{2nd} on the other ten tracks in this task.

\end{abstract}

\section{Introduction}



Named Entity Recognition (NER) is a core natural language processing (NLP) task, which aims at finding entities and recognizing their type in a text sequence.
In practical and open-domain settings, it is difficult for machines to process complex and ambiguous named entities \cite{DBLP:journals/corr/AshwiniC14}.
For example, ``\emph{On the Beach}'' is the title of a movie but cannot be recognized easily by present NER systems.
This issue may become even more serious in multilingual or code-mixed settings \cite{DBLP:conf/sigir/FetahuFRM21}.
However, it has not received sufficient attention from the research community.
To alleviate the issue, SemEval-2022 Task 11 \cite{multiconer-report} formulates this task which focuses on detecting semantically ambiguous and complex entities in short and low-context settings for 11 languages.

One of the classic approaches to solving this problem is to integrate external entity knowledge or gazetteers into neural architectures \cite{DBLP:conf/acl/LiuYL19,DBLP:conf/acl/RijhwaniZNC20,DBLP:conf/naacl/MengFRM21}. 
Typically, the two representations respectively from a language model like BERT \cite{DBLP:conf/naacl/DevlinCLT19} and a gazetteer network like BiLSTM \cite{DBLP:journals/neco/HochreiterS97}  are combined as one merged embedding, which is further fed into a NER classifier such as a conditional random field (CRF) \cite{DBLP:conf/icml/LaffertyMP01}.
However, there is a sense of ``gap'' between the two networks.
The gazetteer network has no explicit semantic learning goal itself, which means it is just a more complex but almost isolated embedding layer for gazetteer information and cannot obtain the true meaning of NER tags actively.
Almost no semantic information can be gained by the classic gazetteer network.

\begin{figure*}[ht]
\includegraphics[width=\textwidth]{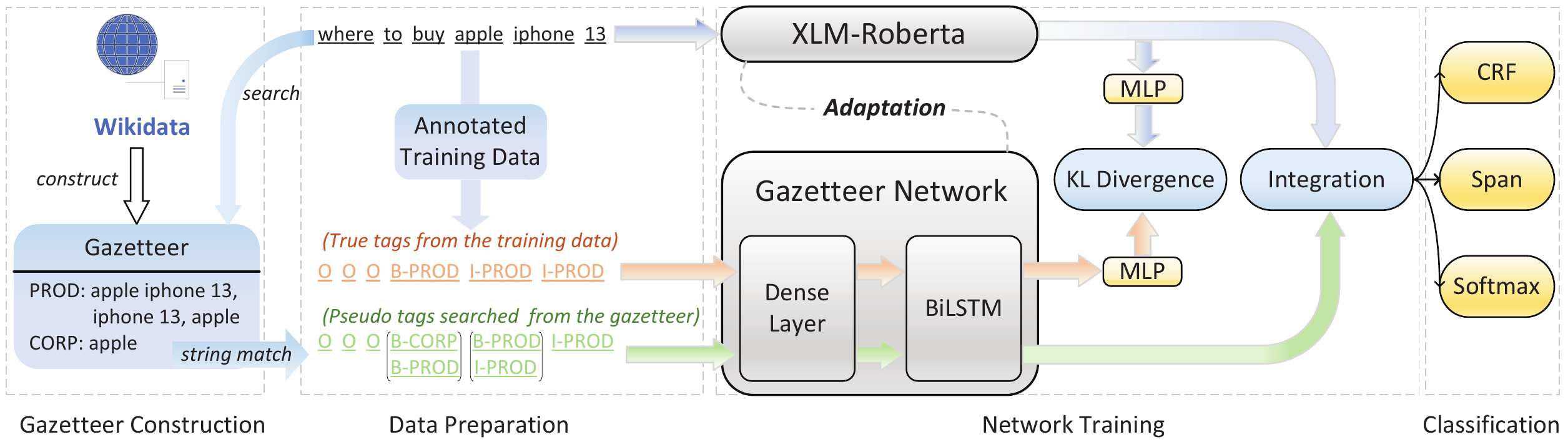}
\caption{The overall structure of the proposed system.}  \label{fig1}

\end{figure*}

To effectively connect the two networks, a {\FULLMODELNAME{}} (\MODELNAME{}) is proposed.
The {\MODELNAME{}} adopts a two-stage training strategy to adapt the gazetteer network to the language model.
During the first training stage, the parameters of a language model are fixed.
Then a sentence and its annotation are fed into the two networks separately.
The two outputs are adapted by minimizing the KL divergence between them.
This training process helps the gazetteer network truly understand the meaning of NER tags by transferring semantic information from the pre-trained language model to the gazetteer network, as the randomly initialized gazetteer network is gradually adapted to the pre-trained language model during the training.
In the second stage, with maintaining the training process above, a gazetteer built from Wikidata is applied to generate pseudo annotations searched by string matching.
A sentence and the corresponding pseudo annotation are then fed into the two pre-trained networks separately.
Finally, integration methods like the concatenation or weighted summation are utilized on the two output representations for classifying.

The proposed method achieves great improvements on the validation set \cite{multiconer-data} of SemEval-2022 Task 11 compared to baseline models and ordinary NER systems with gazetteers.
Ensemble models are used for all thirteen tracks in the final test phase, and our system officially ranked \textbf{1st} on three tracks (Chinese, Code-mixed and Bangla), and \textbf{2nd} on the other ten tracks among nearly 50 teams overall.
The outstanding performance demonstrates the effectiveness of our method.
Fine-grained results show that our method significantly improves scores of difficult labels like ``CREATIVE-WORK'' and ``PRODUCT'', which is the key challenge of this task. To facilitate the reproduction of our results, the code is available at \url{https://github.com/Mckysse/GAIN}.

\section{Task Description}\label{sect2}
SemEval-2022 Task 11 challenges participants to develop complex NER systems for 11 languages (English, Spanish, Dutch, Russian, Turkish, Korean, Farsi, German, Chinese, Hindi, and Bangla), focusing on recognizing semantically ambiguous and complex entities in short and low-context settings \cite{multiconer-report}.
Each language constitutes a single track, while Multilingual and Code-mixed are added as Track 12 and 13.
The task adopts the WNUT 2017 \cite{DBLP:conf/aclnut/DerczynskiNEL17} taxonomy entity types: PERSON (PER for short, names of people), LOCATION (LOC, locations/physical facilities), CORPORATION (CORP, corporations and businesses), GROUPS (GRP, all other groups), PRODUCT (PROD, consumer products), and CREATIVE-WORK (CW, movie/song/book/etc. titles).
The task also aims at testing the domain adaption capability of the systems by adding additional test sets on questions and short search queries.

For each language, a training set with 15300 samples and a validation set with 800 samples are provided. 
For the Code-mixed track, 1500 training samples and 500 validation samples are provided. 
The test data for each track have instances between 150K+ and 500K+ \cite{multiconer-data}.

\section{System Description}
This study focuses on making better use of the external entity knowledge.
To describe our system clearly, in this section, we first introduce three basic mainstream NER systems used.
Then we show the process of constructing a gazetteer with Wikidata, and how the gazetteer representation is generated and utilized.
Finally, we illustrate the {\FULLMODELNAME{}} (\MODELNAME{}).
The overall structure of the proposed system is shown in Figure~\ref{fig1}.

\subsection{Basic NER Systems}\label{sect: backend}
We mainly use the XLM-RoBERTa large \cite{DBLP:conf/acl/ConneauKGCWGGOZ20} as the pre-trained language model, which is a widely used encoder.
Generated by feeding a sentence into the encoder, the representation is then projected to a 13-dimension embedding corresponding to 13 BIO-tags (eg. B-PER, I-PER, O, ...) through a linear transformation.
Three mainstream NER backend classifiers are adopted: Softmax \cite{DBLP:conf/naacl/DevlinCLT19} and CRF \cite{DBLP:journals/corr/HuangXY15} are classic sequential labeling methods that predict the tag of each token, and Span \cite{DBLP:conf/acl/YuBP20} is a segment-based method that predicts the start and the end of an entity separately.

\begin{table*}[]
\begin{center}
\begin{tabular}{l|c|cccccc|c}
\Xhline{1.5pt}
      \multicolumn{2}{c|}{}   & \multicolumn{6}{c|}{Coverage Rate per Label} &                                                                                                                                                                    \\\hline
 Language           & Total Num.                  & \multicolumn{1}{c}{LOC} & \multicolumn{1}{c}{CW} & \multicolumn{1}{c}{PER} & \multicolumn{1}{c}{GRP} & \multicolumn{1}{c}{PROD} & \multicolumn{1}{c|}{CORP} & \multicolumn{1}{c}{Average} \\\Xhline{1pt}
BN                 &73,468                      & 46.52\%                 & 20.21\%                & 7.93\%                  & 19.58\%                 & 10.61\%                  & 14.62\%                  & 19.91\%                     \\\hline
DE                 &901,635                      & 84.34\%                 & 93.62\%                & 19.05\%                 & 87.70\%                 & 47.72\%                  & 94.73\%                  & 71.19\%                     \\\hline
EN                 & 1,032,955                     & 84.26\%                 & 79.30\%                & 18.29\%                 & 86.79\%                 & 47.25\%                  & 94.04\%                  & 68.32\%                     \\\hline
ES                 & 961,535                     & 83.31\%                 & 83.08\%                & 21.03\%                 & 67.32\%                 & 27.51\%                  & 77.81\%                  & 60.01\%                     \\\hline
FA                 & 633,348                     & 87.31\%                 & 77.54\%                & 90.19\%                 & 65.70\%                 & 42.47\%                  & 70.88\%                  & 72.35\%                     \\\hline
HI                 &82,669                      & 46.87\%                 & 13.99\%                & 15.25\%                 & 16.36\%                 & 12.06\%                  & 13.95\%                  & 19.75\%                     \\\hline
KO                 &358,510                      & 58.64\%                 & 79.94\%                & 84.01\%                 & 74.40\%                 & 44.27\%                  & 63.51\%                  & 67.46\%                     \\\hline
NL                 &701,839                      & 79.69\%                 & 72.54\%                & 23.93\%                 & 68.35\%                 & 26.33\%                  & 72.20\%                  & 57.17\%                     \\\hline
RU                 & 495,585                     & 57.17\%                 & 72.19\%                & 9.75\%                  & 41.46\%                 & 28.29\%                  & 57.20\%                  & 44.34\%                     \\\hline
TR                 & 407,901                     & 84.15\%                 & 79.88\%                & 88.54\%                 & 58.29\%                 & 44.26\%                  & 76.75\%                  & 71.98\%                     \\\hline
ZH                 &  513,704                    & 73.42\%                 & 62.84\%                & 19.37\%                 & 81.54\%                 & 34.67\%                  & 71.95\%                  & 57.30\%                     \\\hline
MIX                &  4,434,100                    & 91.71\%                 & 93.44\%                & 44.30\%                 & 90.58\%                 & 55.22\%                  & 96.55\%                  & 78.63\%                     \\\hline
MULTI              &  4,434,100                    & 73.19\%                 & 69.71\%                & 44.49\%                 & 58.65\%                 & 31.44\%                  & 62.00\%                  & 56.58\%                     \\\Xhline{1pt}
\multicolumn{2}{c|}{Average coverage rate} & 73.12\%                 & 69.10\%                & 37.39\%                 & 62.82\%                 & 34.78\%                  & 66.63\%                  & \textbf{57.31\%}              \\      
\Xhline{1.5pt}
\end{tabular}
\end{center}
\caption{The metrics of our gazetteer in detail. The Total Num. column means the accurate number of entries in the gazetteer for each track. Numbers with \% denote the coverage rates to entities in the training and validation set.}\label{tabx}
\end{table*}

\subsection{The Gazetteer}\label{sect: gazetteer}
It's difficult to process complex and ambiguous entities only relying on the language model itself \cite{DBLP:journals/corr/AshwiniC14}. 
To integrate external entity knowledge, we first need to build a large gazetteer matching the taxonomy, then we have to consider how to fuse the gazetteer information with the semantic information from the language model.

\subsubsection{Construction}
Our gazetteer is built based on Wikidata.
Wikidata is a free and open knowledge base.
Every entity of Wikidata has a page consisting of a label, several aliases, descriptions, and one or more entity types.
The entity type annotated by Wikidata is the key to constructing a gazetteer.
For example, ``\emph{apple}'' can be annotated as a kind of fruit or a well-known high-tech corporation in America.
Thus, according to WNUT 2017 taxonomy \cite{DBLP:conf/aclnut/DerczynskiNEL17} used in this competition, the word ``\emph{apple}'' is given both PROD and CORP labels.

To construct a gazetteer fit to the data of this task, firstly every entity of the training set is searched in Wikidata. Then all the entity types returned are mapped to the NER taxonomy with 6 labels.
Next, all Wikidata entities stored in these entity types can be added to the 6 labels gazetteer separately.
Of course, there is a lot of entities that cannot be searched, especially in some languages such as BN and HI.
Also, the elementary gazetteer has plenty of noise.
By measuring the number and the coverage rate of each language on each label, the mapping relationships are adjusted manually.
In the end, a multilingual gazetteer is obtained that contains entities from 70K to 1M for each language.
The gazetteer approximately has a coverage rate of 57 percent on entities in the training and validation set.
Basic information about our gazetteer is shown in Table ~\ref{tabx}.
``Coverage Rate'' is calculated as the number of entities both appeared in the official data and in our gazetteer divided by the total number of entities in the official data.

\begin{table}[t]
    \centering
    \setlength{\tabcolsep}{1.5mm}
    \begin{tabular}{l|ccccc}
    \Xhline{1.5pt}
         \footnotesize{Words} &\footnotesize{O} &\footnotesize{B-CORP} &\footnotesize{I-CORP} &\footnotesize{B-PROD} &\footnotesize{I-PROD}  \\\hline\hline
         where &1 &0 &0 &0 &0 \\\hline
         to &1 &0 &0 &0 &0 \\\hline
         buy &1 &0 &0 &0 &0 \\\hline
         apple &0 &1 &0 &1 &0 \\\hline
         iphone &0 &0 &0 &1 &1 \\\hline
         13 &0 &0 &0 &0 &1 \\\Xhline{1.5pt}
    \end{tabular}
    \caption{Example of the one-hot representation for a searched sentence. The rest 8 labels are all zero.}
    \label{tab2}
\end{table}

\subsubsection{Application}\label{usegaze}
To apply the gazetteer to a sentence, firstly a search tree is constructed for string matching.
Once a sentence is fed into the search tree, a maximum length matching algorithm will be conducted, and a 13-dimension one-hot vector for each token will be generated.
Take the sentence ``\emph{where to buy apple iphone 13}'' for example.
By string matching with the gazetteer, ``\emph{apple iphone 13}'', ``\emph{iphone 13}'' and ``\emph{apple}'' are found in the PROD gazetteer, while ``\emph{apple}'' is also found in the CORP gazetteer.
Then a 13-dimension one-hot vector will be generated for every word as shown in Table~\ref{tab2}.

Denote one sentence as \(\mathbf{w}=(w_1,w_2,...w_N)\) where \(N\) is the length of the sentence and \(w_i\) is the \(i^{th}\) word.
By feeding \(\mathbf{w}\) into the encoder such as the XLM-RoBERTa large, a semantic representation \(\mathbf{e}\in \mathbb{R}^{N\times D}\) is obtained, where D is the hidden size.
At the same time, the one-hot vector generated from the search tree is fed into a gazetteer network consisting of a dense layer and a BiLSTM.
To match the hidden size of the language model, the output embedding \(\mathbf{g}\) has the same size with \(\mathbf{e}\).

Two common ways are used to integrate \(\mathbf{e}\) and \(\mathbf{g}\).
One way is to concatenate them on each token, another way is to get the weighted summation of them by setting a trainable parameter \(\lambda\in \mathbb{R}^{N\times D}\). The final representation is fed into the backend classifier for supervised NER training.

\subsection{Gazetteer-Adapted Integration Network}\label{GAIN}
Through the analysis in Section ~\ref{gradient}, it is found that only conducting the normal training process above is not enough. 
Since the encoder XLM-RoBERTa large and the gazetteer network BiLSTM are almost isolating each other during the whole training, almost no semantic information can be gained explicitly by the classic gazetteer network.

To address this problem, the {\MODELNAME{}} method is proposed as a two-stage training strategy.
In the first stage, the adaptation between the two networks is conducted.
Take the sentence \(\mathbf{w} = \) \{\emph{where to buy apple iphone 13}\} for example.
Assuming the correct tags are \({\rm T} = \)\{O,O,O,B-PROD,I-PROD,I-PROD\}, an one-hot vector is constructed just based on the true tags.
A gazetteer representation \(\mathbf{g}_{r}\in \mathbb{R}^{N\times D}\) is obtained after passing the vector through the gazetteer network.
Then the parameters of the language model are fixed, and the sentence \(\mathbf{w} \) is fed into it to get a semantic representation \(\mathbf{e}\).
\(\{\mathbf{g}_{r}, \mathbf{e}\}\) are projected to \(\{\mathbf{g}^{t}_{r},\mathbf{e}^{t}\} \in \mathbb{R}^{N\times 13}\) by two separate linear layers, where the semantic meaning is transferred to the tags meaning as a kind of logits distributions.
The adaptation is implemented by the designed loss \(\emph{L}_{1}\):

\begin{equation}\label{eq1}
\emph{L}_{1}(\mathbf{w})={\rm KL}(sg(\mathbf{g}^{t}_{r})||\mathbf{e}^{t})+{\rm KL}(sg(\mathbf{e}^{t})||\mathbf{g}^{t}_{r})
\end{equation}

\noindent where \({\rm KL}(\cdot)\) is the KL divergence calculation and \(sg(\cdot)\) operation is used to stop back-propagating gradients, which
is also employed in \citet{DBLP:conf/acl/JiangHCLGZ20,DBLP:journals/corr/abs-2004-08994}.
The loss \(\emph{L}_{1}\) is the symmetrical Kullback-Leibler divergence, encouraging the distributions \(\mathbf{g}^{t}_{r}\) and \(\mathbf{e}^{t}\) to
agree with each other.
Thus, the gazetteer network will understand the real meaning of the NER tags and gain semantic information transferred from the language model.

In the second stage, all the parameters are trained with a gazetteer.
As illustrated in Section \ref{usegaze}, a gazetteer representation \(\mathbf{g}\) is generated from the search tree and the gazetteer network BiLSTM. Next, an ordinary fusion method is applied to \(\mathbf{g}\) and \(\mathbf{e}\) to get an integration representation, which is then fed into the backend classifier to compute a conventional loss with true tags \({\rm T}\). This supervised training goal is implemented by the loss \(\emph{L}_{2}\):

\begin{equation}
\emph{L}_{2}(\mathbf{w}) = {\rm Classifier}(f(\mathbf{g},\mathbf{e}),{\rm T})
\end{equation}
\noindent where \(f(\cdot)\) denotes ordinary integration methods like concatenation or weighted summation. \({\rm Classifier}(\cdot)\) represents one of the three mainstream backend classifiers mentioned in Section \ref{sect: backend}. During the whole second-stage training, a multitask learning goal is conducted shown as:

\begin{equation}
\emph{L}_{3}(\mathbf{w}) = \alpha\emph{L}_{1}(\mathbf{w}) + \emph{L}_{2}(\mathbf{w})
\end{equation}

\noindent where \(\alpha\) is a hyperparameter that is manually set for different fusion and backend methods.


\section{Data Preparation}\label{sec::dataaug}

For the basic training set provided officially, an entity replacement strategy is adopted using our own gazetteer to construct a double data-augmented set. 
This part of data is called ``data-wiki'', which mainly consists of rich-context sentences.

In order to improve the performance of our models on low-context instances, a set of annotated sentences are generated from the MS-MARCO QnA corpus (V2.1) \cite{DBLP:conf/nips/NguyenRSGTMD16} and the ORCAS dataset \cite{DBLP:conf/cikm/CraswellCMYB20}, which are mentioned in \citet{DBLP:conf/naacl/MengFRM21}.
Our trained models and existing NER systems (e.g., spaCy) are applied to identify entities in these corpora, and only templates identically recognized by all models are reserved.
Finally, 3753 English templates for MS-MARCO and 13806 English templates for ORCAS are obtained.
After slotting the templates by our own gazetteer and translating them to the other 10 languages, we get approximately 16K annotated low-context sentences for each language.
This part of data is called ``data-query''.

\begin{table*}[]
    \centering
    \begin{tabular}{l|c}
    \Xhline{1.5pt}
    Parameter & Value\\\hline
    Hidden size for language models & 1024 for large, 768 for base \\
    Learning rate for language models & 2e-5 for large, 1e-5 for base \\
    Learning rate for gazetteer networks & 2e-4 for large, 1e-4 for base\\
    Learning rate for the CRF layer & 2e-3 for large, 1e-3 for base\\
    First-stage training epochs & 5\\
    Second-stage training epochs & 20\\
    Batchsize & 32\\
    Dropout rate & 0.1\\
    \(\alpha\) for the second stage training & 5 for Softmax and Span, 100 for CRF\\
    Optimizer & AdamW\\
    Activation function & Relu
\\\Xhline{1.5pt}
    \end{tabular}
    \caption{Hyperparameters for our system. ``large'' means the 24-layer transformer model, and ``base'' denotes the 12-layer transformer model.}
    \label{tabz}
\end{table*}

\begin{table}[]
    \centering
    \setlength{\tabcolsep}{1.5mm}
    \begin{tabular}{l|c}
    \Xhline{1.5pt}
    Model Name & Lang\\\hline\hline
    XLM-R large \cite{DBLP:conf/acl/ConneauKGCWGGOZ20} & Multi\\\hline
    chinese-roberta-wwm \cite{DBLP:journals/taslp/CuiCLQY21} & ZH\\\hline
    luke-large \cite{DBLP:conf/emnlp/YamadaASTM20} & EN\\\hline
    klue-roberta \cite{DBLP:journals/corr/abs-2105-09680} & KO\\\hline
    bert-base-turkish \cite{stefan_schweter_2020_3770924} &TR \\\hline
    bert-fa-base \cite{DBLP:journals/npl/FarahaniGFM21} &FA
 \\\Xhline{1.5pt}
    \end{tabular}
    \caption{Pre-trained language models used.}
    \label{tab3}
\end{table}

A special operation is conducted for the Code-mixed track, because only 2000 annotated instances are provided officially.
The multilingual function in Wikidata is utilized, as every entity in the Wiki page has several expressions in other languages.
For every sentence in ``data-wiki'' and ``data-query'', the entities inside are randomly replaced with their translations recorded by Wikidata.
In this way, a set of annotated code-mixed data are built.

\section{Experiments}

\subsection{Encoder Selection}
Preliminary experiments show that it's important to start with advanced pre-trained language models for further improvements.
More than a dozen models are evaluated on the development set, and the language models listed in Table~\ref{tab3} are adopted eventually.
For some tracks, the XLM-R large and the monolingual model are both used for ensemble.
For the other tracks without monolingual models in the corresponding language, just the XLM-R large is used.
All the resources can be found on the HuggingFace Page \cite{DBLP:journals/corr/abs-1910-03771}.

\begin{table}
\begin{center}
\setlength{\tabcolsep}{1.5mm}
\begin{tabular}{lccc}
\Xhline{1.5pt}
Track     & Team Num.           & F1        & Rank              \\ \Xhline{1pt}

English (EN) & 30    & 	0.8547        & 2         \\ \hline
Spanish (ES) & 18    & 0.8544        & 2              \\ \hline
Dutch (NL)   & 15        & 0.8767    & 2   \\ \hline
Russian (RU)    & 14         & 0.8382         & 2            \\ \hline
Turkish (TR)  & 15 & 0.8552       & 2        \\ \hline
Korean (KO)    & 17          & 0.8636        & 2       \\ \hline
Farsi (FA)    & 15         & 0.8705      & 2        \\ \hline
German (DE)  & 16        & 	0.8905       & 2    \\ \hline
Chinese (ZH)   & 21         & 0.8169         &1       \\ \hline
Hindi (HI)   & 17        & 0.8464       & 2        \\ \hline
Bangla (BN)    & 18          & 0.8424         &1         \\ \hline
Multilingual  & 26        & 	0.853        & 2          \\ \hline
Code-mixed  & 26        & 0.929       &1       \\ \Xhline{1.5pt}
\end{tabular}
\end{center}
\caption{Official rankings of our system. ``F1'' denotes the macro-F1 on the test set.}\label{tab1}

\end{table}

\subsection{Training Details}
A lot of models have been trained with the {\MODELNAME{}} method using different classifiers, different integration methods, different chosen language models.
Hyperparameter settings are shown in Table~\ref{tabz}.
Sample codes are already available at \url{https://github.com/Mckysse/GAIN}.

A 5-fold cross-validation training strategy is also applied in the evaluation.
The prepared data ``data-wiki'' and ``data-query'' are split into five pieces, each one is used as the validation set, while the other four pieces are used as the training set.
After obtaining the five best models by this strategy, the logits of them (for Softmax and Span models) are averaged to integrate them as an aggregated model. CRF models are just voted averagely.

\begin{table*}[]
\footnotesize
    \centering
    \setlength{\tabcolsep}{1.8mm}{
    \begin{tabular}{clcccccccccccc}
    \Xhline{1pt}
\multicolumn{1}{l}{Strategy}                               & Classifier & BN             & DE             & EN             & ES             & FA             & HI             & KO             & NL             & RU             & TR             & ZH             & MIX            \\\hline
\multirow{3}{*}{A}                                  & CRF       & 0.771          & 0.886          & 0.846          & 0.834          & 0.78           & 0.771          & 0.813          & 0.878          & 0.802          & 0.835          & 0.866          & 0.654          \\
                                                           & Softmax   & 0.763          & 0.879          & 0.849          & 0.836          & 0.783          & 0.767          & 0.811          & 0.871          & 0.792          & 0.835          & 0.862          & 0.652          \\
                                                           & Span      & 0.793          & 0.896          & 0.853          & 0.845          & 0.806          & 0.802          & 0.831          & 0.879          & 0.809          & 0.839          & 0.884          & 0.696          \\\hline
\multirow{3}{*}{B}                      & CRF       & 0.816          & 0.906          & 0.865          & 0.857          & 0.821          & 0.8            & 0.853          & 0.888          & 0.817          & 0.865          & 0.908          & 0.788          \\
                                                           & Softmax   & 0.799          & 0.901          & 0.865          & 0.859          & 0.824          & 0.796          & 0.851          & 0.879          & 0.815          & 0.864          & 0.901          & 0.786          \\
                                                           & Span      & 0.811          & 0.917          & 0.871          & 0.857          & 0.818          & 0.825          & 0.864          & 0.887          & 0.82           & 0.858          & 0.906          & 0.792          \\\hline
\multirow{3}{*}{C}  & CRF       & 0.841          & 0.943          & 0.891          & 0.87           & 0.835          & 0.831          & 0.871          & 0.902          & 0.829          & 0.884          & 0.913          & 0.833          \\
\multicolumn{1}{l}{}                                       & Softmax   & 0.829          & 0.931          & 0.888          & 0.872          & 0.839          & 0.822          & 0.868          & 0.897          & 0.831          & 0.882          & 0.909          & 0.835          \\
\multicolumn{1}{l}{}                                       & Span      & 0.832          & 0.935          & 0.892          & 0.874          & 0.836          & 0.837          & 0.879          & 0.901          & 0.836          & 0.872          & 0.912          & 0.823          \\\hline
\multicolumn{2}{l}{weighted token-vote}                                   & \textbf{0.864} & \textbf{0.955} & \textbf{0.922} & \textbf{0.892} & \textbf{0.855} & \textbf{0.853} & \textbf{0.899} & \textbf{0.916} & \textbf{0.843} & \textbf{0.903} & \textbf{0.922} & \textbf{0.865} \\\Xhline{1pt}
\end{tabular}}
    \caption{All macro-F1 scores on the validation set. Only scores of the concatenation integration method are listed due to limited spaces. Strategy A denotes baseline systems mentioned in Section~\ref{sect: backend}, B denotes ordinary integration method with the gazetteer mentioned in Section~\ref{sect: gazetteer}, and C denotes the {\MODELNAME{}} method mentioned in Section~\ref{GAIN}. ``weighted token-vote'' represents the ensemble of all our models including those with the weighted summation integration method not listed, and achieves the best performance on the validation set.}
    \label{tab4}
\end{table*}

\begin{table*}[]
\footnotesize
    \centering
        \setlength{\tabcolsep}{1.7mm}{
\begin{tabular}{l|ccccccccccccc}
\Xhline{1pt}
Coverage Rate & BN    & DE    & EN    & ES    & FA    & HI    & KO    & NL    & RU    & TR    & ZH    & MIX   & avg   \\\Xhline{0.75pt}
0        & 0.784 & 0.897 & 0.856 & 0.847 & 0.8   & 0.775 & 0.839  & 0.892 & 0.806 & 0.855 & 0.863 & 0.662 & 0.823 \\\hline
30\%     & 0.791 & 0.898 & 0.861 & 0.845 & 0.804 & 0.799 & 0.84 & 0.893 & 0.814 & 0.856 & 0.872 & 0.694 & 0.831 \\\hline
50\%     & 0.858 & 0.901 & 0.867 & 0.844 & 0.807 & 0.871 & 0.866 & 0.897 & 0.814 & 0.861 & 0.903 & 0.709 & 0.85  \\\hline
70\%     & 0.891 & 0.907 & 0.868 & 0.854 & 0.811 & 0.899 & 0.894 & 0.901 & 0.82  & 0.869 & 0.904 & 0.732 & 0.863 \\\hline
100\%    & 0.974 & 0.973 & 0.942 & 0.92  & 0.903 & 0.978 & 0.938 & 0.94  & 0.91  & 0.91  & 0.964 & 0.914 & 0.934\\\Xhline{1pt}
\end{tabular}}
    \caption{F1 scores of the gradient coverage rate trial. ``Coverage Rate'' means the number of entities in official data also found in our gazetteer / the number of entities in official data. ``avg'' denotes the average result of all tracks.}
    \label{tab5}
\end{table*}

Finally, the predictions of our best models in different methods are token-voted by setting a weight for each track.
The weight is manually set referring to all scores on the validation set.

\subsection{Official Results}
Our team participate in all the 13 tracks and the results in the test phase are listed in Table~\ref{tab1}. We ranked \textbf{1st} on three tracks and \textbf{2nd} on ten tracks. 

\section{Analysis}
\label{sec:analysis}

\subsection{Effectiveness of The Gazetteer}
To explore the effectiveness of the proposed {\MODELNAME{}} methods, a large number of trials are conducted on the official data mentioned in Section~\ref{sect2}.
All scores under the concatenation integration setting on the validation set are listed in Table~\ref{tab4}.
Significant improvements are gained by the gazetteer and the {\MODELNAME{}} method on all tracks.
The results demonstrate that the gazetteer plays a pivotal role in processing complex entities.

\subsection{Coverage Rate Trial}\label{gradient}

Our gazetteer can only reach approximately 57\%
coverage rate over the entities in the official data. 
Intuitively, the higher the coverage rate over the entities reaches, the better the performance can achieve. 
We carry out a toy experiment to explore this conjecture.
Since entities from the official training and development set can be extracted, we can control how many of these entities appear in our gazetteer to modulate the coverage rate.

As shown in Table~\ref{tab5}, not as expected, scores on many tracks don't improve in step with the increase of the coverage rate when it does not reach 100\%.
This situation mostly happens in languages that already have good scores, like DE and EN.
To explore the reason why the gazetteers under 100\% coverage rate don't work, we check the weight \(\lambda\) mentioned in Section~\ref{usegaze}.
It's surprising to find the \(\lambda\) is nearly zero when the coverage rate is not 100\%, which indicates that our models almost don't use the gazetteer information.
Further fine-grained tests find that only when the coverage rate exceeds the basic accuracy of the model, the \(\lambda\) starts to have a non-zero value.
An empirical conclusion can be drawn that the gazetteer network and the language model almost process information separately, and the final integration module simply selects the better one for classifying.
Thus, this study starts to figure out an explicit way to adapt the two networks, and the {\MODELNAME{}} method is designed.

\begin{table*}[]
\footnotesize
    \centering
    \setlength{\tabcolsep}{0.85mm}{
    \begin{tabular}{c|l|ccccccccccc}
    \Xhline{1.5pt}
\multicolumn{1}{l|}{Domain} & Metrics\textbackslash{}Lang & bn     & de     & en     & es     & fa     & hi     & ko     & nl     & ru     & tr     & zh     \\\hline
\multirow{13}{*}{overall}  & macro@F1                    & 0.8424 & 0.8905 & 0.8547 & 0.8544 & 0.8705 & 0.8464 & 0.8636 & 0.8767 & 0.8382 & 0.8552 & 0.8169 \\
                           & macro@P                     & 0.8584 & 0.8988 & 0.8641 & 0.8664 & 0.8816 & 0.8600 & 0.8739 & 0.8856 & 0.8485 & 0.8662 & 0.8394 \\
                           & macro@R                     & 0.8343 & 0.8835 & 0.8467 & 0.8439 & 0.8610 & 0.8392 & 0.8556 & 0.8686 & 0.8289 & 0.8470 & 0.8007 \\
                           & ALLTRUE                     & 135634 & 271979 & 272922 & 265748 & 192194 & 144940 & 217281 & 265942 & 249458 & 149369 & 173183 \\
                           & ALLPRED                     & 133563 & 269699 & 269658 & 261020 & 189525 & 143238 & 215304 & 262626 & 245126 & 148025 & 169983 \\
                           & ALLRECALLED                 & 123479 & 253694 & 244826 & 238000 & 172403 & 132392 & 197725 & 242799 & 219455 & 135212 & 152918 \\
                           & MD@F1                       & 0.9174 & 0.9367 & 0.9025 & 0.9036 & 0.9033 & 0.9188 & 0.9142 & 0.9187 & 0.8874 & 0.9093 & 0.8912 \\
                           & F1@LOC                      & 0.8339 & 0.8900 & 0.8452 & 0.8606 & 0.8853 & 0.8719 & 0.8789 & 0.8793 & 0.8258 & 0.8427 & 0.8575 \\
                           & F1@PER                      & 0.8724 & 0.9333 & 0.9315 & 0.9224 & 0.9234 & 0.9160 & 0.9100 & 0.9268 & 0.8634 & 0.9096 & 0.8021 \\
                           & F1@PROD                     & 0.7539 & 0.8803 & 0.8221 & 0.7885 & 0.8324 & 0.7694 & 0.8225 & 0.8421 & 0.8166 & 0.8348 & 0.8121 \\
                           & F1@GRP                      & 0.8956 & 0.8777 & 0.8622 & 0.8585 & 0.8670 & 0.8642 & 0.8683 & 0.8891 & 0.8329 & 0.8479 & 0.7966 \\
                           & F1@CORP                     & 0.8921 & 0.8961 & 0.8761 & 0.8799 & 0.8827 & 0.8674 & 0.8767 & 0.8949 & 0.8671 & 0.8722 & 0.8422 \\
                           & F1@CW                       & 0.8067 & 0.8655 & 0.7909 & 0.8165 & 0.8325 & 0.7894 & 0.8253 & 0.8278 & 0.8237 & 0.8240 & 0.7911 \\\hline
\multirow{13}{*}{lowner}   & macro@F1                    & 0.8656 & 0.9541 & 0.9209 & 0.8910 & 0.8529 & 0.8518 & 0.9037 & 0.9143 & 0.8420 & 0.9005 & 0.9101 \\
                           & macro@P                     & 0.8732 & 0.9580 & 0.9283 & 0.9015 & 0.8613 & 0.8649 & 0.9064 & 0.9214 & 0.8522 & 0.9032 & 0.9276 \\
                           & macro@R                     & 0.8582 & 0.9503 & 0.9138 & 0.8808 & 0.8449 & 0.8393 & 0.9011 & 0.9074 & 0.8322 & 0.8980 & 0.8942 \\
                           & ALLTRUE                     & 15699  & 151645 & 152604 & 147408 & 72862  & 25836  & 97937  & 146090 & 129475 & 29940  & 52767  \\
                           & ALLPRED                     & 15436  & 150512 & 150411 & 144103 & 71764  & 25073  & 97614  & 144131 & 126514 & 29840  & 51610  \\
                           & ALLRECALLED                 & 14426  & 147460 & 142921 & 133408 & 62804  & 22985  & 89837  & 136200 & 109732 & 27358  & 49264  \\
                           & MD@F1                       & 0.9267 & 0.9760 & 0.9433 & 0.9153 & 0.8685 & 0.9030 & 0.9188 & 0.9386 & 0.8573 & 0.9153 & 0.9440 \\
                           & F1@LOC                      & 0.9018 & 0.9657 & 0.9441 & 0.9034 & 0.8701 & 0.8961 & 0.9131 & 0.9457 & 0.8130 & 0.9038 & 0.9544 \\
                           & F1@PER                      & 0.9413 & 0.9762 & 0.9717 & 0.9572 & 0.8935 & 0.9273 & 0.9199 & 0.9621 & 0.8387 & 0.9238 & 0.9457 \\
                           & F1@PROD                     & 0.8112 & 0.9308 & 0.8395 & 0.7974 & 0.7677 & 0.7884 & 0.8812 & 0.8449 & 0.8202 & 0.8375 & 0.8849 \\
                           & F1@GRP                      & 0.8647 & 0.9519 & 0.9332 & 0.8989 & 0.8728 & 0.8691 & 0.9000 & 0.9237 & 0.8443 & 0.9287 & 0.8445 \\
                           & F1@CORP                     & 0.8616 & 0.9607 & 0.9409 & 0.9311 & 0.8691 & 0.8460 & 0.9230 & 0.9219 & 0.8924 & 0.9260 & 0.9247 \\
                           & F1@CW                       & 0.8132 & 0.9396 & 0.8963 & 0.8580 & 0.8439 & 0.7842 & 0.8851 & 0.8876 & 0.8436 & 0.8835 & 0.9063 \\\hline
\multirow{13}{*}{orcas}    & macro@F1                    & 0.8237 & 0.7933 & 0.7559 & 0.7968 & 0.8720 & 0.8316 & 0.8182 & 0.8164 & 0.8223 & 0.8317 & 0.7564 \\
                           & macro@P                     & 0.8420 & 0.8050 & 0.7678 & 0.8099 & 0.8820 & 0.8466 & 0.8323 & 0.8263 & 0.8316 & 0.8422 & 0.7796 \\
                           & macro@R                     & 0.8178 & 0.7880 & 0.7505 & 0.7917 & 0.8657 & 0.8287 & 0.8115 & 0.8109 & 0.8171 & 0.8270 & 0.7473 \\
                           & ALLTRUE                     & 101225 & 101171 & 101197 & 101220 & 101149 & 101116 & 101027 & 101160 & 101153 & 101050 & 101179 \\
                           & ALLPRED                     & 99529  & 100039 & 100222 & 99970  & 99741  & 100211 & 99465  & 100097 & 100006 & 100135 & 99159  \\
                           & ALLRECALLED                 & 91027  & 87729  & 84475  & 88455  & 92126  & 91762  & 90263  & 89032  & 91767  & 90378  & 85483  \\
                           & MD@F1                       & 0.9069 & 0.8720 & 0.8388 & 0.8793 & 0.9172 & 0.9116 & 0.9004 & 0.8848 & 0.9124 & 0.8985 & 0.8534 \\
                           & F1@LOC                      & 0.7682 & 0.7374 & 0.6745 & 0.7621 & 0.8613 & 0.8138 & 0.7985 & 0.7365 & 0.7909 & 0.7742 & 0.7410 \\
                           & F1@PER                      & 0.8445 & 0.8486 & 0.8511 & 0.8606 & 0.9342 & 0.8990 & 0.8881 & 0.8684 & 0.8776 & 0.8963 & 0.7327 \\
                           & F1@PROD                     & 0.7393 & 0.8217 & 0.8021 & 0.7773 & 0.8673 & 0.7608 & 0.7756 & 0.8395 & 0.8152 & 0.8343 & 0.7637 \\
                           & F1@GRP                      & 0.9027 & 0.7711 & 0.7633 & 0.8089 & 0.8628 & 0.8633 & 0.8418 & 0.8453 & 0.8195 & 0.8262 & 0.7923 \\
                           & F1@CORP                     & 0.8951 & 0.8167 & 0.7989 & 0.8202 & 0.8896 & 0.8707 & 0.8393 & 0.8653 & 0.8389 & 0.8596 & 0.8000 \\
                           & F1@CW                       & 0.7923 & 0.7641 & 0.6456 & 0.7519 & 0.8169 & 0.7820 & 0.7661 & 0.7437 & 0.7914 & 0.7997 & 0.7084 \\\hline
\multirow{13}{*}{msq}      & macro@F1                    & 0.7512 & 0.8683 & 0.8174 & 0.8510 & 0.9175 & 0.9094 & 0.8819 & 0.8695 & 0.8591 & 0.8783 & 0.8537 \\
                           & macro@P                     & 0.7721 & 0.8706 & 0.8010 & 0.8449 & 0.9245 & 0.9189 & 0.8845 & 0.8630 & 0.8537 & 0.8764 & 0.8594 \\
                           & macro@R                     & 0.7344 & 0.8665 & 0.8366 & 0.8595 & 0.9111 & 0.9014 & 0.8797 & 0.8787 & 0.8671 & 0.8810 & 0.8484 \\
                           & ALLTRUE                     & 18710  & 19163  & 19121  & 17120  & 18183  & 17988  & 18317  & 18692  & 18830  & 18379  & 19237  \\
                           & ALLPRED                     & 18598  & 19148  & 19025  & 16947  & 18020  & 17954  & 18225  & 18398  & 18606  & 18050  & 19214  \\
                           & ALLRECALLED                 & 18026  & 18505  & 17430  & 16137  & 17473  & 17645  & 17625  & 17567  & 17956  & 17476  & 18171  \\
                           & MD@F1                       & 0.9663 & 0.9660 & 0.9139 & 0.9474 & 0.9653 & 0.9819 & 0.9646 & 0.9473 & 0.9593 & 0.9595 & 0.9452 \\
                           & F1@LOC                      & 0.9372 & 0.9334 & 0.8548 & 0.9113 & 0.9553 & 0.9685 & 0.9421 & 0.9055 & 0.9209 & 0.9245 & 0.9296 \\
                           & F1@PER                      & 0.9276 & 0.9335 & 0.9366 & 0.9252 & 0.9771 & 0.9688 & 0.9525 & 0.9368 & 0.9399 & 0.9428 & 0.8762 \\
                           & F1@PROD                     & 0.7890 & 0.8211 & 0.7734 & 0.7810 & 0.8848 & 0.8435 & 0.7858 & 0.8129 & 0.7255 & 0.8267 & 0.7886 \\
                           & F1@GRP                      & 0.0000 & 0.7797 & 0.7273 & 0.7527 & 0.8771 & 0.8518 & 0.8357 & 0.8079 & 0.8063 & 0.7994 & 0.7868 \\
                           & F1@CORP                     & 0.9365 & 0.8603 & 0.8214 & 0.8574 & 0.9110 & 0.9231 & 0.8811 & 0.8937 & 0.8708 & 0.8797 & 0.8665 \\
                           & F1@CW                       & 0.9171 & 0.8817 & 0.7909 & 0.8784 & 0.8999 & 0.9007 & 0.8939 & 0.8605 & 0.8909 & 0.8967 & 0.8745 \\\Xhline{1.5pt}
\end{tabular}}
    \caption{All detailed results of the official test set on monolingual tracks.}
    \label{taball}
\end{table*}

\begin{table*}[]
\footnotesize
    \centering
    \setlength{\tabcolsep}{0.6mm}{
    \begin{tabular}{l|cccccccccc}
    \Xhline{1pt}
Track\textbackslash{}Metrics & macro@F1 & macro@P & macro@R & MD@F1  & F1@LOC & F1@PER & F1@PROD & F1@GRP & F1@CORP & F1@CW  \\\hline
Code-mixed                     & 0.9290   & 0.9321  & 0.9261  & 0.9662 & 0.9314 & 0.9461 & 0.9164  & 0.9247 & 0.9463  & 0.9093 \\
Multilingual                 & 0.8530   & 0.8605  & 0.8489  & 0.9210 & 0.8681 & 0.9076 & 0.8105  & 0.8152 & 0.8786  & 0.8381 \\\Xhline{1pt}
\end{tabular}}
    \caption{Results of the official test set on the Code-mixed and Multilingual tracks.}
    \label{tab2m}
\end{table*}

\subsection{Detailed Results on the Test Set}
As shown in Table~\ref{taball} and Table~\ref{tab2m}, we display all the detailed results on all tracks during the test phase for further analyses in-depth.

As shown in the tables, the proposed GAIN method significantly improves the performance of the language model on recognizing entities of hard labels like ``CW'' and ``PROD''.
Besides, results in the domain of ``orcas'' and ``msq'' indicate the effectiveness of the proposed data-augment strategy mentioned in Section~\ref{sec::dataaug}.

\subsection{Average-Logits Experiments}

This section explains why we choose to average logits of softmax-based models (for Softmax and Span models) for integrating them as an aggregated model, rather than averagely token-vote.
Also, a 5-fold cross-validation training is conducted with the official training data on the basic Softmax method.
Without loss of generality, BN, EN, FA, and ZH are chosen to represent different language families.
Results on the official validation set are shown in Table~\ref{taby}.
It is empirically demonstrated that average-logits for the softmax-based model ensemble is better than average-token-vote in most situations.

\begin{table}[h]
    \centering
    \setlength{\tabcolsep}{1.5mm}{
    \begin{tabular}{l|cccc}
    \Xhline{1.5pt}
Data\textbackslash{}Lang & bn    & en    & fa    & zh    \\\hline
1 fold                        & 0.795 & 0.871 & 0.814 & 0.874 \\
2 fold                        & 0.799 & 0.873 & 0.8   & 0.871 \\
3 fold                        & 0.801 & 0.876 & 0.795 & 0.889 \\
4 fold                        & 0.798 & 0.881 & 0.791 & 0.881 \\
5 fold                        & 0.789 & 0.869 & 0.8   & 0.873 \\\hline
avg                      & 0.796 & 0.874 & 0.8   & 0.878 \\
avg-token-vote           & 0.813 & 0.887 & 0.815 & 0.898 \\
avg-logits               & \textbf{0.815} & \textbf{0.89}  & \textbf{0.817} & \textbf{0.903}\\\Xhline{1.5pt}
\end{tabular}}
    \caption{Results of the 5-fold cross-validation trial. ``avg'' denotes the average results of 5 models' scores. ``avg-token-vote'' represents the averagely token-vote process. ``avg-logits'' means average logits of 5 models are fed into the backend softmax layer for classifying, which achieves the best performance on all 4 languages.}
    \label{taby}
\end{table}

\subsection{About KL}
This section illustrates why the GAIN method adopts the KL divergence rather than the MMSE loss.
Softmax-based classification usually chooses the position of the maximum in all dimensions to be the predicted tag.
Because the softmax module and logits-softmax module exist in the KL calculation, the 13-dimension vector can be considered as a kind of logits distribution with the tags meaning.
Thus, the distribution, or the internal relationship between all logits of tokens in a sentence, is the most important thing to pay attention to, rather than the amplitude of all logits.
With the KL divergence, two distributions are adapted to each other without considering the amplitude.

A toy trial has been conducted to confirm whether the two 13-dimension vectors from the two networks have the same amplitude.
The result shows that they are not always in the same ratio relations for different inputs.
Compared to the MMSE loss which also calculates relating to the absolute amplitude difference, the KL divergence is better since it only focuses on the distributions of the two outputs, without being disturbed by the inconsistent relationship of the amplitude.

\section{Conclusion}
This paper presents the implementation of the USTC-NELSLIP system submitted to the SemEval-2022 Task 11 MultiCoNER.
The {\MODELNAME{}} method is proposed to adapt the gazetteer network to the language model, and achieves great improvements on the complex NER task.
Some construction methods for gazetteers and code-mixed data augmentation are also provided.
In future works, we will keep exploring effective ways to integrate gazetteer networks with encoders to make better use of the external entity knowledge.





\end{document}